\documentclass[runningheads]{llncs}
\usepackage{graphicx}
\usepackage{algorithm}
\usepackage{algpseudocode}
\usepackage{physics}
\usepackage{adjustbox} 
\usepackage{soul}
\usepackage{color}
\usepackage{xcolor}
\usepackage{enumitem}
\setlist[itemize]{label=$\cdot$}
\usepackage{amssymb}
\usepackage[colorlinks,citecolor=blue,urlcolor=blue, linkcolor=blue,hypertexnames=True]{hyperref}

\begin{document}
%

\title{TTA-OOD: Test-time Augmentation for Improving Out-of-Distribution Detection in Gastrointestinal Vision}

\titlerunning{TTA-OOD in Gastrointestinal Vision}
%
\author{Sandesh Pokhrel\inst{1}\thanks{Equal Contribution} \and
Sanjay Bhandari\inst{1}\textsuperscript{*} \and
Eduard Vazquez\inst{2} \and
Tryphon Lambrou \inst{4} \and
Prashnna Gyawali \inst{3} \and
Binod Bhattarai \inst{1,4}
}
\authorrunning{Pokhrel and Bhandari et al.}

\institute{Nepal Applied Mathematics and Informatics Institute for research(NAAMII), Lalitpur, Nepal \and
Fogsphere(Redev AI Ltd.), 64 Southwark Bridge Rd, SE1 0AS, London, UK \and
 West Virginia University, USA\and
School of Natural and Computing Sciences, University of Aberdeen, Aberdeen, UK\\
\email{\{binod.bhattarai\}@abdn.ac.uk}
}

%
\maketitle              
\begin{abstract}
Deep learning has significantly advanced the field of gastrointestinal vision, enhancing disease diagnosis capabilities. One major challenge in automating diagnosis within gastrointestinal settings is the detection of abnormal cases in endoscopic images. Due to the sparsity of data, this process of distinguishing normal from abnormal cases has faced significant challenges, particularly with rare and unseen conditions. To address this issue, we frame abnormality detection as an out-of-distribution (OOD) detection problem. In this setup, a model trained on In-Distribution (ID) data, which represents a healthy GI tract, can accurately identify healthy cases, while abnormalities are detected as OOD, regardless of their class. We introduce a test-time augmentation segment into the OOD detection pipeline, which enhances the distinction between ID and OOD examples, thereby improving the effectiveness of existing OOD methods with the same model. This augmentation shifts the pixel space, which translates into a more distinct semantic representation for OOD examples compared to ID examples. We evaluated our method against existing state-of-the-art OOD scores, showing improvements with test-time augmentation over the baseline approach.

\end{abstract}

\section{Introduction}

Out of all the reported diseases worldwide, diseases with a digestive etiology have burdened the field of medicine with over seven billion incidents reported in 2019 alone \cite{DigestiveDiseaseBurden}. Owing to technological advancements, the fatality rates for most diseases have decreased; however, the number of deaths from gastrointestinal diseases increased from 2000 to 2019 \cite{DigestiveDiseaseBurden}. Invasive endoscopic procedures are vital for diagnosing various diseases, aiding in the detection of abnormalities and guiding necessary treatments. Recent advancements in deep learning, particularly Convolutional Neural Networks (CNNs), show great promise in automating these procedures. The rise of portable endoscopy methods, such as capsule endoscopy, has amplified the demand for automated diagnostic tools. CNNs have been effective in detecting diseases like gastric cancer \cite{gastriccancerAICNN}, polyps, ulcerative colitis, and esophagitis \cite{svmcnngastrodisease,gitractensemblestacking,dlendodisease}. They have also been used to classify anatomical landmarks in endoscopic images. While these models perform well in recognizing normal landmarks, their accuracy decreases with abnormalities and rare cases, leading to trust issues \cite{oodmedicalapplications}. 
\noindent \textbf{OOD Methods} help models detect anomalies, improving their deployability in real-world scenarios without much modifications. Researchers utilize logits, probability, gradient, and feature space information for generalized out-of-distribution detection. Scoring methods in OODs condense this information to identify in-distribution (ID) or out-of-distribution (OOD) samples.  OOD detection methods range from approaches like MSP \cite{msp} and Entropy \cite{entropy}, which use softmax probabilities, to techniques like maximum logit \cite{maxlogit} and energy score \cite{energyscore}. Advanced methods such as ODIN \cite{Odin}, ViM \cite{vim} and Mahanalobis \cite{mahalanobis} incorporate additional information like gradients and feature space distances. KNN-OOD \cite{knn-ood} takes a non-parametric approach, leveraging nearest neighbour distances in feature space to identify out-of-distribution samples.
\noindent \textbf{Test-time augmentations (TTAs)} have been demonstrated to effecteively enhance the performance of various deep learning models across different domains. Moshkov et al. (2020) \cite{moshkov2020test} showed that simple TTAs, like rotation and flipping, improve cell segmentation accuracy in microscopy images. Wang et al. (2019) \cite{wang2019automatic} found TTAs boost brain tumor segmentation performance. He et al. (2022) \cite{ttaanamoly} demonstrated TTAs' ability to distinguish between in-distribution and out-distribution samples for OOD detection.

OOD detection in the medical domain is crucial as it only requires healthy examples to identify abnormalities. MOOD \cite{MOOD} highlights the importance of flagging OODs in medical imaging to catch erroneous predictions early. Arnau et al. \cite{selfsupendo} proposed a self-supervised approach using ODIN as an OOD scoring metric to separate anomalies from healthy abdomen sections. Another work \cite{endoood} uses ViM for OOD detection with mixup and long-tail ID data calibration techniques. Mehta et al. \cite{oodskinlesion} employed mixup in long-tail data for skin lesion detection, classifying new and emerging skin abnormalities. Motivated by these works, we study the effect of perturbations in image space and how it translates to semantic representation and whether any tell-tale signs of OOD data can be extracted from the shift in the representation.
A new direction of research is emerging in gastrovision as well where models are trained solely on normal anatomical findings for landmark classification, and any deviation in endoscopic images is treated as an abnormality \cite{endoood,selfsupendo}. However, there is a lack of studies exploring this approach in diverse settings with different Out-of-Distribution (OOD) scoring methods and how these scores perform under augmentation. We address this problem by formulating diseases as abnormalities in gastrointestinal images. We propose a supervised approach to the classification of anatomical structures in gastrovision, while these scoring metrics assist us in discerning healthy regions from unhealthy ones.
Our contributions can be summarised as:

\begin{itemize}
\renewcommand\labelitemi{{\normalfont \bfseries $\bullet$}} 
    \item We conceptualize abnormalities in gastrovision as Out-of-Distribution (OOD) instances, eliminating the necessity for models to be trained on abnormal images to detect them.
    \item We enhance existing OOD detection methods by integrating test time augmentations and assess their performance in perturbed scenarios.
    \item Our approach is simple, model-agnostic, and OOD score-independent, making it versatile and easy to implement across various model architectures and compatible with any OOD scoring method.
\end{itemize}

We evaluated our method using different classifier backbone architectures (ViT and ResNet) on the KvasirV2 dataset \cite{kvasir}, showcasing its effectiveness.

\section{Method}
We study how test time augmentations can improve OOD detection methods in identifying out-of-distribution instances, particularly in gastrointestinal settings. While existing methods effectively distinguish in-distribution from OOD examples, they struggle with near-OOD cases that have close semantic similarity. We explore how test time augmentations on robust models create drift for test samples, enhancing OOD detection performance. Our method integrates easily with existing architectures and OOD detection approaches without changing their processes.

\begin{figure}[t]
\includegraphics[width=\linewidth]{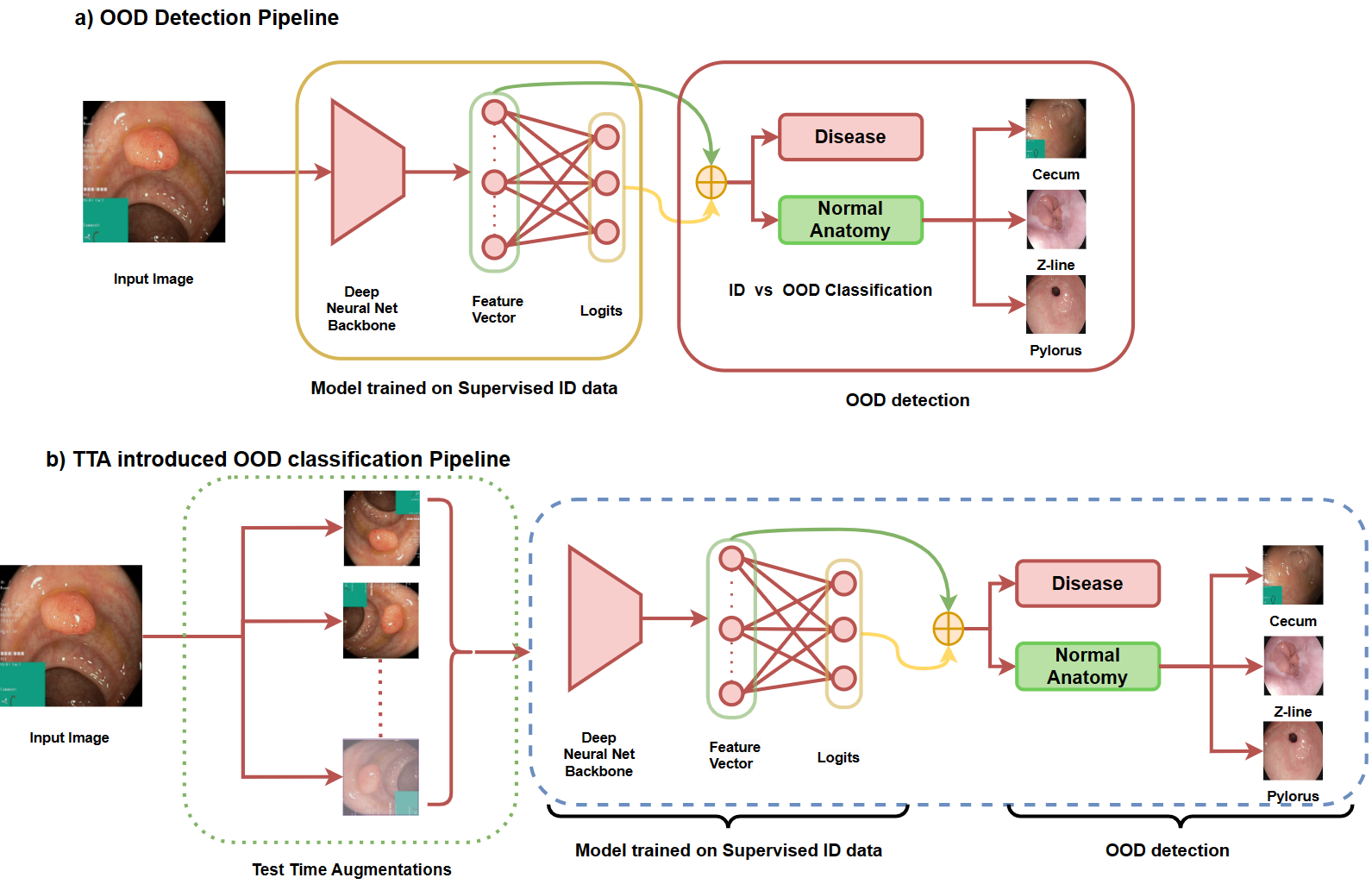}
\centering
\caption{\textbf{a)} OOD based classification pipeline which separates ID (normal anatomy) and OOD (Disease) data based on feature, logit or gradient information. \textbf{b)} TTA based OOD detection pipeline which works on images with individual or composite augmentations.} \label{fig1}
\end{figure}

\noindent \textbf{Problem Formulation:} We focus on supervised multi-class classification where, $X$ is input space, $Y = \{1, 2, \ldots, C\}$ is label space and $\text{D}_{in}= {(x_i, y_i)}^{n}_{i=1}$ is training set (i.i.d. from $P_{X, Y}$), composed of healthy gastrointestinal images (in-distribution data). A neural network $f: {X} \rightarrow \mathbb{R}^{|Y|}$ is trained on ID data using cross-entropy loss to classify healthy classes. The penultimate layer $\varphi: X \rightarrow \mathbb{R}^m$ produces feature vector $\varphi(x)$ for feature-based OOD detection. Logits-based methods use the final linear layer output for OOD detection.

\noindent \textbf{Data Augmentations:}
A set of pre-defined transformations is employed to generate augmented versions of each test-time sample. These transformations can be broadly categorized into:

\noindent \textbf{Individual Augmentations (IA):} These base operations modify a single aspect of the data. Examples include horizontal flip (H), vertical flip (V), and color jittering (CJ). For ID samples, these pixel-level transformations often result in only minor changes to the semantic representations learned by the model. However, for OOD samples, the same augmentations can lead to more significant shifts in the feature space representation, as the model has not encountered these types of images during training. This discrepancy in the model's response to augmented ID and OOD samples can be leveraged to enhance the performance of OOD detection methods.

\noindent \textbf{Composite Augmentations (CA):} These combine multiple individual augmentations to create more diverse variations. A composite augmentation can be represented as the sequential application of individual transformations (e.g. H $\rightarrow$ V $\rightarrow$CJ, H $\rightarrow$ V). In contrast to individual augmentation, composite augmentations can induce more complex and diverse perturbations to the input data which leads to more significant shift in the feature space. This desired perturbation cause more pronounced drifts for OOD samples than ID samples, which results in larger separations between ID and OOD data representations and ultimately enhancing the OOD detection performance. This idea is also empirically verified by Table \ref{tab:TTA on resnet}, Table \ref{tab:TTA on ViT} and Figure \ref{fig2}(a), where composite augmentation seems to be more effective in enhancing the OOD detection algorithm's ability to differentiate between in-distribution (ID) and out-of-distribution (OOD) samples, highlighting the utility of composite augmentations in improving OOD detection methods.
The equation below formalizes the generation of an augmented sample (x'):
\begin{equation}
x' = T(x)
\end{equation}
where, $x$ is original input sample and $T$ is transformation, which can be either an individual augmentation ($IA_i$) or a composite augmentation ($C(IA_j, ..., IA_k)$).

\noindent \textbf{OOD Score Calculation:}
Augmented samples are processed through a neural network trained on in-distribution (ID) data. The model's output is used to compute an OOD score for test samples, with the specific calculation method varying based on model architecture and desired properties. This approach exploits the model's response to augmented data to identify deviations from the ID data manifold, potentially indicating OOD samples.
Random transformations introduce controlled variations in test-time data, simulating drift and making samples appear more OOD to the model. We hypothesize that a robust model will show minimal feature space distinction for pixel-space perturbations in ID inputs, while OOD samples, being unfamiliar, will exhibit greater differences in semantic understanding.

Following this, the decision for whether a test image is OOD or not can be formulated as
\begin{equation} \label{eq4}
\begin{split}
g(x) = & \left\{
\begin{array}{ll}
\textit{OOD}, & \text{if } \text{score} \geq \lambda \\
\textit{ID}, & \text{otherwise}
\end{array}
\right. \\
\end{split}
\end{equation}
where $\lambda$ is the threshold that is decided ensuring that the validation set retains at least a given true-positive rate.

\section{Experiments}

\subsection{Datasets and Implementation Details}
We utilized the Kvasirv2 \cite{kvasir} multi-class endoscopy dataset annotated which is verified by experienced endoscopist and is designed for a range of tasks including  Gastrointestinal Disease Detection.

\noindent \textbf{Dataset:}
The \textbf{Kvasirv2}\cite{kvasir} dataset consists 8 categories of upper and lower GI tract anatomical landmarks, most common pathological findings, or endoscopic procedures within the gastrointestinal (GI) tract. The three anatomical landmarks provided in this dataset are the Z-line, Pylorus, and Cecum. Pathological finding in the dataset include Esophagitis(ESO), Polyps(POL), and Ulcerative colitis(UC). In addition to normal and anatomical findings this dataset also consist of images related to polyp removal, the ``dyed and lifted polyp"(DLP) class and the ``dyed resection margins"(DRM).
For our experimental settings, the anatomical landmarks are our classification task classes(ID-data) that represent healthy GI regions. The remaining five classes are marked as OOD as they constitute of some form of abnormality that you would see in the abdomen. 
For model training, we selected 2400 in-distribution images(80\%) and used 600 in-distribution images(20\%), combined with 5000 out-of-distribution images, to assess the model's final performance on OOD capabilities.\\
\noindent \textbf{Implementation Details:}
The Resnet-18 model was trained for 20 epochs. The ViT-Small model was also trained for 20 epochs with a patch size of 16. The batch size in training was 32 and we used the Adam optimizer with an initial learning rate of 1×10$^{-4}$ for both models. The models were initialized with Imagenet pre-trained weights for better accuracy. The input image size for both models is resized to 224x224 pixels and the standard cross-entropy loss function was used  to train the models. The models were trained and tested in PyTorch v2.1.0 with an NVIDIA A100 GPU. 

\begin{table}
    \centering
    \caption{Metrics of the optimal model on downstream task (classification of normal landmarks on Kvasirv2 dataset) which is used for OOD detection}
    \begin{tabular}{c c  c c c}
        \hline
        \hline
        \textbf{Model} & \multicolumn{1}{c}{\textbf{Parameters}} & \multicolumn{1}{c}{\textbf{Feat.Dim}} & \multicolumn{1}{c}
        {\textbf{Accuracy}}  \\
        \hline
        \hline
        Resnet-18 & 11M & 512 & 98.57   \\
        ViT-small & 21M & 384 & 99.10  \\
        \hline
    \end{tabular} 
    \label{tab: Downstream_task}
\end{table}

\subsection{Results}
Our study presents enhancing out-of-distribution (OOD) detection methods by employing test time augmentations in gastrointestinal contexts. We assessed the capabilities of established OOD detection techniques,logit-based methods like MSP \cite{msp}, ODIN \cite{Odin}, Energy \cite{energyscore}, Entropy \cite{entropy}, MaxLogit \cite{maxlogit}, feature-based methods like Mahalanobis \cite{mahalanobis}, and method like Vim \cite{vim} which combine both feature as well as logit information and investigated how test time augmentations induce subtle drift in test samples and exploited this phenomenon to improve OOD detection performance of respective methods. The model deemed optimal for downstream classification task was selected as the feature extractor for OOD detection task shown in Table \ref{tab: Downstream_task}.

\noindent \textbf{Quantitative Results:}
AUC and FPR95 are the most commonly used metrics to evaluate the OOD detection performance. AUC measures the area under the Receiver Operator Characteristic curve, with higher values indicating better performance. FPR represents the false positive rate when the true positive rate is 95\%, with smaller values indicating better performance.

\begin{table}[t]
\centering
\caption{Quantitative Comparison of enhancement of different OOD methods with TTA in terms of AUC$\uparrow$ and FPR$\downarrow$ for Resnet-18 on Kvasirv2 dataset. Both values are presented in percentage. Bold characters denote the best improvement for a particular OOD method under respective augmentation and underline represents the improvement over baseline with respective TTA.}
\label{tab:TTA on resnet}
\small
\begin{adjustbox}{max width=\textwidth}
\begin{tabular}{ccccccccccccc}
\hline
Augmentations & \multicolumn{2}{c}{MSP} & \multicolumn{2}{c}{Odin} & \multicolumn{2}{c}{Energy} & \multicolumn{2}{c}{MaxLogit} & \multicolumn{2}{c}{Mahalanobis} & \multicolumn{2}{c}{ViM} \\
\hline
& AUC & FPR & AUC & FPR & AUC & FPR & AUC & FPR & AUC & FPR & AUC & FPR \\
\hline
None & 87.57 & 33.06 & 86.95 & 40.48 & 86.00 & 43.28 & 86.04 & 43.04 & 88.58 & 33.06 & 89.65 & 29.86 \\
Hflip & \underline{87.97} & 34.80 & \underline{87.20} & \underline{37.88} & \underline{86.38} & \underline{40.18} & \underline{86.43} & \underline{39.88} & \underline{88.76} & 33.34 & \underline{89.84} & \underline{27.7} \\
Vflip & \underline{87.72} & 34.42 & 86.86 & \underline{40.42} & \underline{86.10} & \underline{41.54} & \underline{86.15} & \underline{41.22} & \underline{88.88} & \underline{31.98} & 89.5 & \textbf{26.42} \\
Color Jitter & \underline{87.66} & 34.1 & 86.83 & \underline{39.56} & 85.43 & 46.76 & 85.93 & \underline{41.24} & 88.31 & 35.7 & \underline{89.78} & \underline{28.48} \\
Hflip + Vflip & \underline{87.98} & 34.9 & \underline{87.08} & 41.12 & \underline{86.42} & \underline{42.00} & \textbf{86.46} & \underline{41.76} & \textbf{89.63} & \textbf{31.28 }& \textbf{90.37} & \underline{26.7} \\
Hflip + Color jit & \textbf{87.99} & 33.28 & \textbf{87.23}& \textbf{37.66} & \textbf{86.39} & \textbf{38.86} & \underline{86.32} & \textbf{38.44} & 88.25 & \underline{32.62} & 89.59 & \underline{28.7} \\
Vflip + Color jit & 87.57 & \textbf{32.46} & 86.46 & \underline{40.34} & 85.72 & \underline{40.22} & 85.89 & \underline{40.86} & \underline{88.7} & \underline{33.02} & 89.42 & \underline{27.92} \\
H + V + Color jit & \underline{87.66} & 35.34 & 86.78 & 41.32 & 86.22 & 43.2 & \underline{86.12} & 44.8 & \underline{89.2} & \underline{32.78} & \underline{90.18} & \underline{27.00} \\
\hline
\end{tabular}
\end{adjustbox}
\end{table}

\begin{table}[htbp]
\centering
\caption{Quantitative Comparison of enhancement of different OOD methods with TTA in terms of AUC$\uparrow$ and FPR$\downarrow$ for ViT-small on Kvasirv2 dataset. Bold characters denote the best improvement for a particular method under respective augmentation and underline represents improvement over baseline with respective TTA.}
\label{tab:TTA on ViT}
\small
\begin{adjustbox}{max width=\textwidth}
\begin{tabular}{ccccccccccccc}
\hline
Augmentations & \multicolumn{2}{c}{MSP} & \multicolumn{2}{c}{Odin} & \multicolumn{2}{c}{Energy} & \multicolumn{2}{c}{MaxLogit} & \multicolumn{2}{c}{Mahalanobis} & \multicolumn{2}{c}{ViM} \\
\hline
& AUC & FPR & AUC & FPR & AUC & FPR & AUC & FPR & AUC & FPR & AUC & FPR \\
\hline
None & 85.93 & 39.74 & 83.84 & 42.1 & 83.52 & 43.12 & 83.53 & 43.12 & \textbf{93.06} & 25.66 & 93.70 & 26.34 \\
Hflip & 85.16 & 43.32 & 82.79 & 43.78 & 82.41 & 44.34 & 82.41 & 44.32 & 92.36 & 28.06 & 93.59 & 28.3 \\
Vflip & \underline{87.04} & \underline{36.5} & \underline{85.11} & \underline{40.1} & \underline{84.11} & \textbf{41.76} & 84.11 & \textbf{41.74} & 91.99 & \textbf{22.42} & 93.54 & \underline{22.08} \\
Color Jitter & \underline{86} & \underline{39.32} & \underline{83.92} & \underline{41.86} & 83.51 & 43.48 & 83.51 & \underline{42.5} & 93.00 & 25.92 & \underline{93.74} & \underline{25.48} \\
Hflip + Vflip & \underline{86.35} & \underline{36.58} & \underline{84.24} & \underline{41.3} & 83.37 & 43.82 & 83.37 & 43.82 & 91.38 & \underline{23.52} & 92.83 & \underline{20.92} \\
Hflip + Color Jit  & 85.22 & 43.2 & 82.8 & 43.52 & 82.4 & 44.82 & 82.47 & 43.84 & 92.35 & 27.12 & 93.55 & 29 \\
Vflip + Color Jit  & \textbf{87.14} & \textbf{36.16} & \textbf{85.2}& \textbf{39.92} & \textbf{84.25} & \underline{42.34} & \textbf{84.25} & \underline{42.74} & 91.97 & \underline{21.06} & \textbf{93.86} & \underline{20.98} \\
H + V + Color Jit  & \underline{86.38} & \underline{36.28} & \underline{84.43} & \underline{41.3} & 83.46 & 43.32 & 83.42 & 44.02 & 91.34 & \underline{24.26} & 92.88 & \textbf{20.3} \\
\hline
\end{tabular}
\end{adjustbox}
\end{table}
Tables \ref{tab:TTA on resnet} and \ref{tab:TTA on ViT} show that test-time augmentation significantly improves out-of-distribution (OOD) detection performance. For Resnet-18, ViM achieves the lowest False Positive Rate (FPR) of 26.7\% with Vertical flipping, a 3.16\% improvement over the baseline. For ViT-small, ViM achieves a 20.3\% FPR with Horizontal flipping, Vertical flipping, and Color Jitter, a 6.04\% enhancement. It's evident that the application of test-time augmentations consistently enhances the performance of all OOD detection baseline methods. Achieving a low FPR95 is particularly crucial as it directly contributes to reducing false positives and thereby increasing the trustworthiness of the model's predictions. By incorporating test-time augmentations, we not only strengthen the robustness of the model but also equip it with better generalization capabilities in terms of OOD detection, which are paramount for real-world applications where unseen data instances are common.

\begin{figure}[t]
\includegraphics[width=1\textwidth]{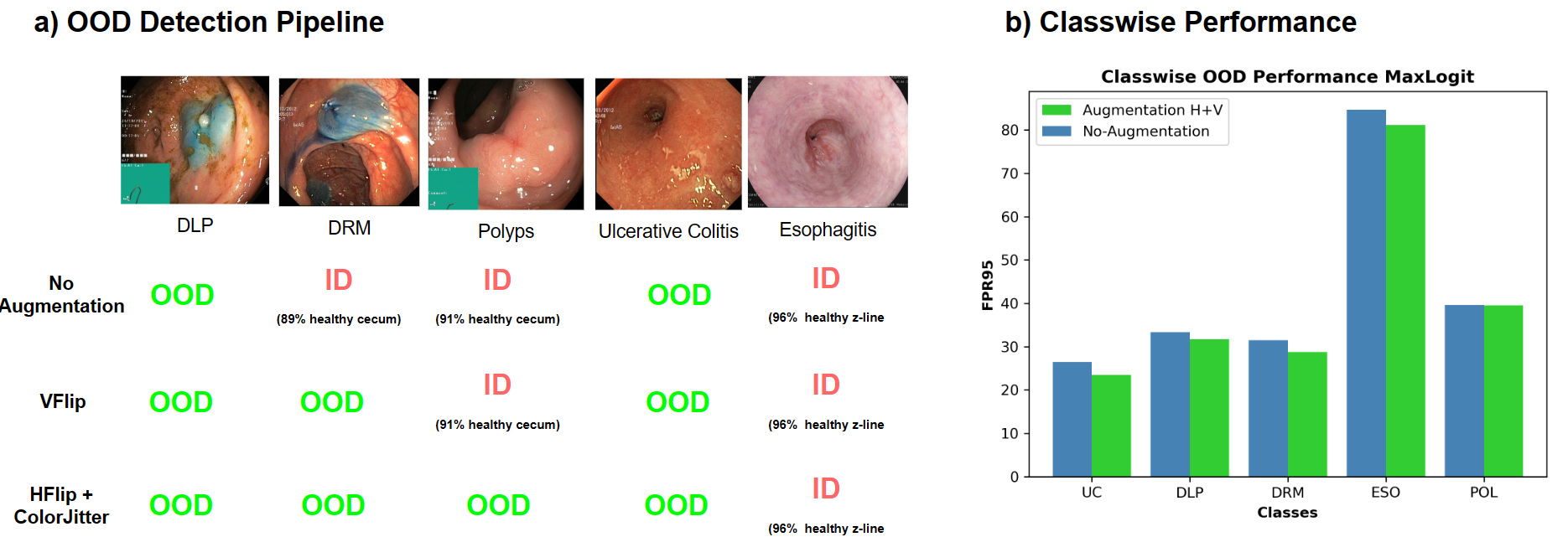}
\caption{\textbf{a)} 
Qualitative comparison of Maxlogit method for Kvasirv2 on Resnet18: OOD examples over-confidently predicted by the corresponding method as healthy ID data (red) and correctly identified as abnormality (green) under respective test time augmentation technique. 
\textbf{b)} 
Performance improvement on Maxlogit method on FPR95$\downarrow$ on separate OOD classes for Kvasirv2 on Resnet18 model.
}
\label{fig2}
\end{figure}

\noindent \textbf{Qualitative Results:}
The qualitative results in Fig \ref{fig2} shows the improvement in performance of MaxLogit method on different OOD classes when subject to augmentation. The effect of different augmentations and their predictions either OOD or misclassified as healthy organs are presented in Fig \ref{fig2}.a. It is evident from these results that Test Time Augmentation introduces drift in representation space allowing existing OOD metrics to distinguish ID and OOD better. Moreover, the drift introduced in representation space is different for different augmentations owing to difference in pixel space perturbation. In general, composite augmentations are stronger and create a larger drift as seen in the predictions. 

\noindent \textbf{Ablation Study:}
Ablation experiments were performed to assess the impact of different augmentation methods on out-of-distribution (OOD) scoring performance. Table \ref{tab:TTA ablations} shows that Horizontal Flip (Hflip), Vertical Flip (Vflip), and Color Jitter improve OOD detection for both logit-based and feature-based methods. Conversely, Equalize and Invert exhibit a detrimental effect, reducing the performance for both methods. In addition, Table \ref{tab:TTA ood scoring} demonstrates that augmenting OOD images causes a drift in representation space, aiding in distinguishing OOD data. The mean OOD score, higher for OOD data, increases with augmentation. ViM shows an average drift of 2.71, raising the mean OOD score from 6.70 to 9.41 with Horizontal and Vertical Flipping. Similar results are observed in logits-based methods like MaxLogit.

\begin{table}[ht]
\centering
\caption{Ablation study for impact of different augmentation method in terms of AUC$\uparrow$ and FPR$\downarrow$ for Resnet-18 on Kvasirv2 dataset.}
\label{tab:TTA ablations}
\small
\begin{adjustbox}{max width=\textwidth}
\begin{tabular}{ccccc}
\hline
Augmentations &  \multicolumn{2}{c}{MaxLogit}  & \multicolumn{2}{c}{ViM} \\
\hline
& AUC & FPR & AUC & FPR \\
\hline
None & 86.04 & 43.04 & 89.65 & 29.86 \\
Hflip & \textbf{86.43} & \textbf{39.88} & \underline{89.84} & \underline{27.7} \\
Vflip & \underline{86.15} & \underline{41.22} & 89.5 & \textbf{26.42} \\
Color Jitter & 85.93 & \underline{41.24} & \underline{89.78} & \underline{28.48} \\
Equalize & \underline{86.43} & 46.88 & \textbf{90.36} & 30.00 \\
Invert & 82.27 & 58.88 & 89.16 & 33.94 \\

\hline
\end{tabular}
\end{adjustbox}
\end{table}

\begin{table}[htb]
\centering
\caption{Ablation study on drift introduced by TTA that enhances the robustness of OOD detection methods. The average OOD score on OOD examples increases indicating that we get larger values for each OOD example in general which resonates with the fact that OOD examples get higher score values than ID data.}
\label{tab:TTA ood scoring}
\small
\begin{adjustbox}{max width=\textwidth}
\begin{tabular}{ccc}
\hline
Augmentations &  \multicolumn{2}{c}{Mean OOD Score} \\
\hline
\hline
& MaxLogit & ViM  \\
\hline
None  & 5.98 & 6.70 \\
Hflip + Vflip & 6.03 & 9.41  \\

\hline
\end{tabular}
\end{adjustbox}
\end{table}

\section{Conclusion}
We reformulate abnormality detection in gastrointestinal images as an out-of-distribution (OOD) problem, enabling the recognition of abnormalities in the GI tract using a model trained solely on normal anatomical findings. To achieve this, we introduce test-time augmentation (TTA), which significantly improves OOD detection performance by augmenting the input data during inference. This approach, when combined with existing OOD detection methods from the OOD literature, enhances their generalization and robustness regarding OOD detection. Overall, our findings highlight the considerable potential of supervised models in OOD detection, serving as a versatile tool for identifying abnormalities in endoscopy images from gastrointestinal settings—an area largely unexplored in medical imaging research.

\newpage
\bibliographystyle{splncs04}
\bibliography{reference.bib}

\end{document}